\def\year{2022}\relax
\documentclass[letterpaper]{article} 
\usepackage{aaai22}  
\usepackage{times}  
\usepackage{helvet}  
\usepackage{courier}  
\usepackage[hyphens]{url}  
\usepackage{graphicx} 
\usepackage{natbib}  
\usepackage{caption} 
\DeclareCaptionStyle{ruled}{labelfont=normalfont,labelsep=colon,strut=off} 
\usepackage{algorithm}
\usepackage{algorithmic}
\usepackage{amsmath}
\usepackage{multirow} 
\usepackage{booktabs}
\usepackage{color}

\usepackage{newfloat}
\usepackage{listings}
\floatstyle{ruled}
\newfloat{listing}{tb}{lst}{}
\floatname{listing}{Listing}
\title{Assessing a Single Image in Reference-Guided Image Synthesis}
\author {
    Jiayi Guo\textsuperscript{\rm 1}, 
     Chaoqun Du\textsuperscript{\rm 1}, 
     Jiangshan Wang\textsuperscript{\rm 2}, 
     Huijuan Huang\textsuperscript{\rm 3}, 
     Pengfei Wan\textsuperscript{\rm 3}, 
     Gao Huang\textsuperscript{\rm 1,4}\footnote{Corresponding author.}
}
\affiliations {
    \textsuperscript{\rm 1}Department of Automation, BNRist, Tsinghua University, Beijing, China\\
    \textsuperscript{\rm 2}Beijing University of Posts and Telecommunications, Beijing, China\\
    \textsuperscript{\rm 3}Y-tech, Kuaishou Technology\\
    \textsuperscript{\rm 4}Beijing Academy of Artificial Intelligence, Beijing, China\\
    \{guo-jy20, dcq20\}@mails.tsinghua.edu.cn, hill@bupt.edu.cn, \{huanghuijuan, wanpengfei\}@kuaishou.com, gaohuang@tsinghua.edu.cn
}

\usepackage{bibentry}

\begin{document}

\maketitle

\begin{abstract}
Assessing the performance of Generative Adversarial Networks (GANs) has been an important topic due to its practical significance. Although several evaluation metrics have been proposed, they generally assess the quality of the \emph{whole} generated image distribution. For Reference-guided Image Synthesis (RIS) tasks, i.e., rendering a source image in the style of another reference image, where assessing the quality of a \emph{single} generated image is crucial, these metrics are not applicable. In this paper, we propose a general learning-based framework, Reference-guided Image Synthesis Assessment (RISA) to quantitatively evaluate the quality of a single generated image. Notably, the training of RISA does not require human annotations. In specific, the training data for RISA are acquired by the intermediate models from the training procedure in RIS, and weakly annotated by the number of models' iterations, based on the positive correlation between image quality and iterations. As this annotation is too coarse as a supervision signal, we introduce two techniques: 1) a pixel-wise interpolation scheme to refine the coarse labels, and 2) multiple binary classifiers to replace a naïve regressor. In addition, an \emph{unsupervised} contrastive loss is introduced to effectively capture the style similarity between a generated image and its reference image. Empirical results on various datasets demonstrate that RISA is highly consistent with human preference and transfers well across models.
\end{abstract}


\begin{figure*}[t]
    \centering
    \includegraphics[scale=0.65]{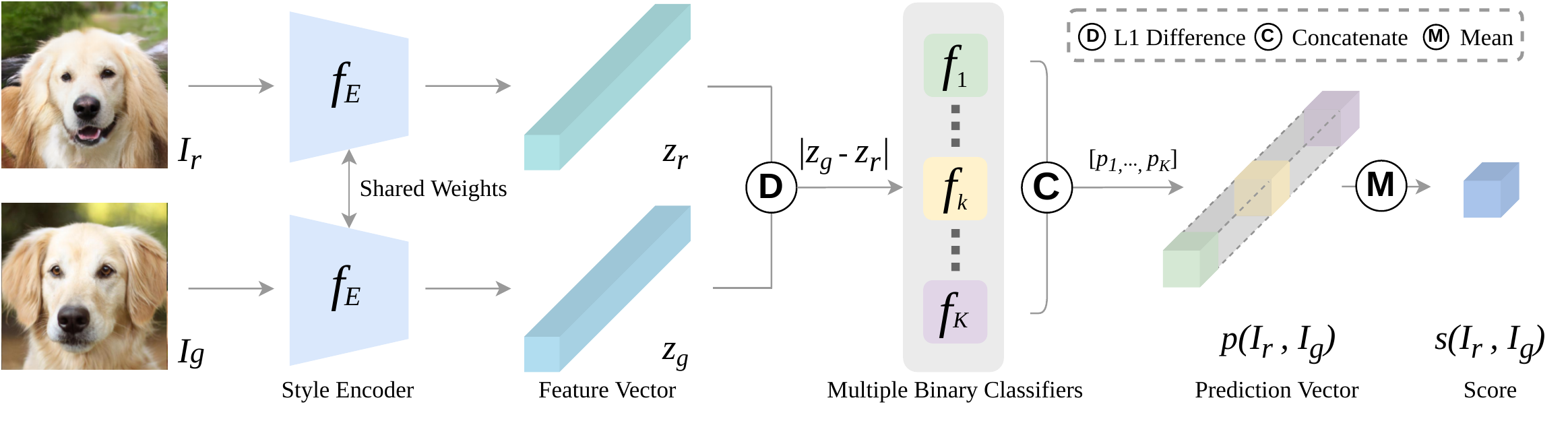}
    \caption{The pipeline of the proposed Reference-guided Image Synthesis Assessment (RISA). RISA consists of a style encoder and multiple binary classifiers. Given a generated image $I_{\rm {g}}$ and its reference image $I_{\rm {r}}$, RISA first utilizes the style encode to extract their style codes $z_{\rm {g}}$ and $z_{\rm {r}}$, respectively. Then the discrepancy (L1 difference) between $z_{\rm {g}}$ and $z_{\rm {r}}$ is calculated and fed into multiple binary classifiers. Finally, quality score is given by averaging the predictions of all classifiers.}
    \label{fig:pipeline}
    \vskip -0.1in
\end{figure*}

\section{Introduction}
Reference-guided Image Synthesis (RIS) aims to utilize Generative Adversarial Networks (GANs) \cite{goodfellow2014generative} to modify the style of a source image to that of a reference image. As described in recent image translation works \cite{lee2018diverse, huang2018multimodal,choi2020stargan}, style refers to the unique appearance of a single image, while the underlying spatial structure is defined as content. Style also coincides with the definition of texture in some other works \cite{park2020swapping}. Nowadays, generative models are widely deployed to provide various RIS services, such as modifying a user's facial features to that of a super star, or changing a building's original appearance to another.

To enhance user experience in RIS applications, it is of great practical signficance to quantitatively evaluate the images generated by GANs. Although several sample-based GAN evaluation metrics have been proposed \cite{xu2018empirical}, e.g., Kernel MMD \cite{gretton2012optimal}, Inception Score (IS) \cite{salimans2016improved}, Mode Score (MS) \cite{che2016mode}, Wasserstein distance and Fréchet Inception Distance (FID) \cite{heusel2017gans}, they mainly focus on assessing the \emph{whole} generated image distribution. In specific, the discrepancy between feature distributions of real and generated images is computed as a quality measure.

However, these metrics are not applicable to evaluate a \emph{single} generated image. In interactive RIS applications, a user may submit one source image (or a pair of source image and reference image) at a time and expects to obtain a satisfying generated image. Unfortunately, due to the notoriously unstable training procedure of GANs, it is challenging to guarantee that each generated image is synthesized with high quality, especially when the source image and the reference image have a large style discrepancy.

Hence, it is important to design an assessment metric for a single image in RIS to improve user experience. Once the quality of each generated image could be effectively assessed, we could simultaneously deploy several different models to generate images for a task and automatically render the image of the highest quality score to users. If all these images are synthesized with low quality, we could refuse to provide any image. However, recent works on single image assessment are either designed to report the average quality score with dozens of images \cite{shaham2019singan} or not able to capture the style similarity between a generated image and its reference \cite{bosse2016deep, talebi2018nima, zhang2018unreasonable, gu2020giqa}.

In this paper, we propose a general learning-based framework, Reference-guided Image Synthesis Assessment (RISA), through which the quality of a single generated image can be effectively assessed. As illustrated in Figure \ref{fig:pipeline}, given a generated image and its reference image, RISA first extracts their style codes via the style encoder. Then the difference of style codes is calculated as the input of multiple binary classifiers. Finally, the quality score is obtained by averaging the predictions of all classifiers.

\begin{figure*}[t] 
    \centering
    \includegraphics[scale=0.4]{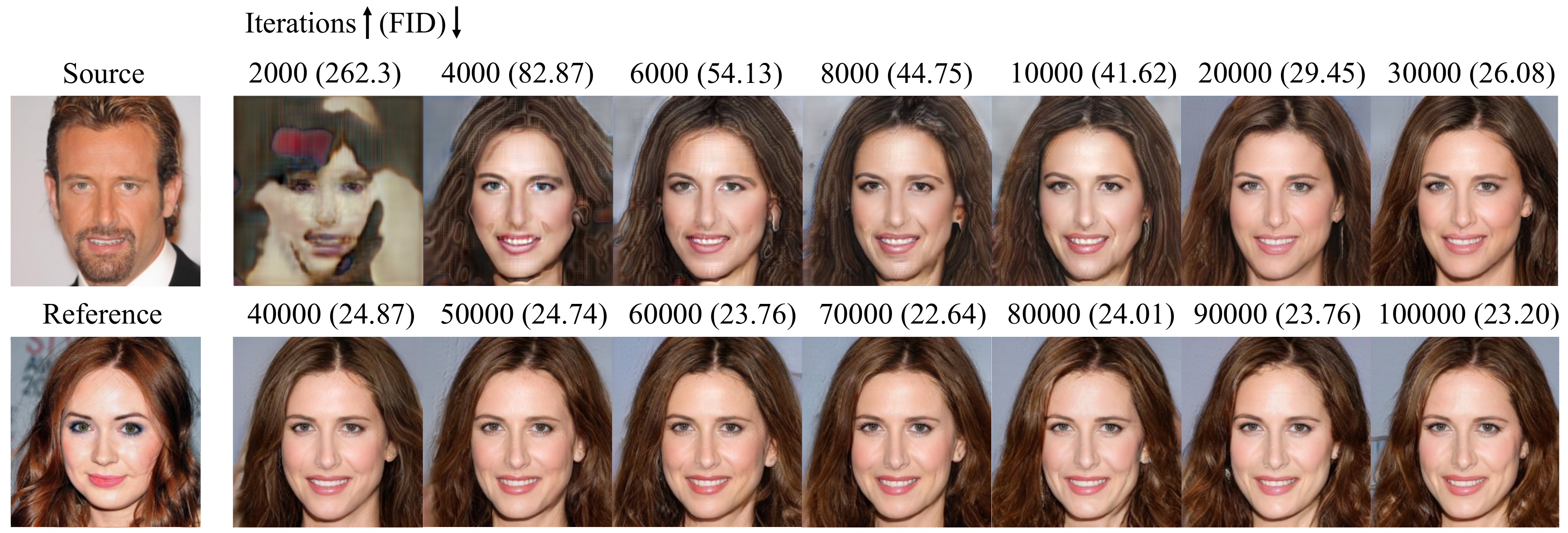}
    \vskip -0.05in
    \caption{Visualizations and Fréchet Inception Distance (FID) variations of StarGAN v2 on CelebA-HQ dataset with the increase of training iterations. The first column shows the source image and the reference image of a specific image synthesis task, while remaining columns are images generated by intermediate models at different training iterations. The parentheses above each generated image gives FID score at corresponding training iterations.}
    \label{fig:iter}
    \vskip -0.1in
\end{figure*}

RISA works in a weakly supervised scheme, i.e., it does not require any human annotations. As illustrated in Figure \ref{fig:iter}, the generated image quality generally increases with training iterations during the GAN training procedure. Therefore, we leverage images generated by intermediate models as the training data for RISA, and consider the number of model's training iterations as a pseudo quality label. Note that a naïve implementation of this idea leads to degenerated solutions, due to that the supervision signal is too coarse, as shown by the results in Table \ref{tab:loss-ablation}. To address this issue, we adopt a pixel-wise interpolation technique to generate images with different quality levels instead of directly utilizing images synthesized by intermediate models during the stable stage. To further suppress the label noise, we deploy multiple binary classifiers rather than a naïve regressor.

Moreover, RISA is optimized by a novel objective, containing 1) a weakly supervised loss to fit the quality label using binary cross entropy, 2) a contrastive loss to effectively capture the style similarity between a generated image and its reference image and 3) a supremum loss to learn the style consistency between two style-preserving augmentation views of a same real image.

We evaluate the effectiveness of RISA on various datasets. Compared with existing single image quality assessment metrics, empirical results  demonstrate that our method achieves higher consistency with human preference,  and transfers well across different models.

\section{Related Work}

\textbf{Reference-guided image synthesis.}  In the context of neural style transfer, reference-guided image synthesis aims to render a source image in the style of a reference image. \citet{gatys2015texture, gatys2016image} utilize feature statistics of deep neural network to capture the style of an image for the first time. \citet{huang2017arbitrary} propose AdaIN to implement arbitrary style transfer. More recently, StarGAN \cite{choi2018stargan} and StarGAN v2 \cite{choi2020stargan} learn mapping between multiple domains with a single generator. MSGAN \cite{mao2019mode} proposes mode seeking regularization to resolve the mode collapse problem. DRIT \cite{lee2018diverse}, MUNIT \cite{huang2018multimodal} and Swapping Autoencoder \cite{park2020swapping} focus on the style and content disentanglement in the feature space.

\noindent \textbf{Image Quality Assessment (IQA).} According to the availability of reference, IQA methods are generally divided into three categories: 
1) \emph{Full-Reference IQA} (FR-IQA) refers to estimating the quality of natural images with references. Widely-used FR-IQA metrics include MS-SSIM \cite{wang2003multiscale}, SSIM \cite{wang2004image}, PSNR \cite{huynh2008scope}, FSIM \cite{zhang2011fsim} and LPIPS \cite{zhang2018unreasonable}.
2) \emph{Reduced-Reference IQA} (RR-IQA) tackles situations where the reference image is not fully accessible. Representative methods are local-harmonic based algorithm \cite{gunawan2003reduced} and grouplet-based algorithm \cite{maalouf2009grouplet}.
3) \emph{No-Reference IQA} (NR-IQA) assesses distorted image without any reference. Early works include support vector regression based methods \cite{moorthy2010two,moorthy2011blind} and probability based methods \cite{mittal2012making}. With the prevalence of deep learning, massive of network architectures \cite{bosse2016deep,liu2017rankiqa,talebi2018nima,lin2018hallucinated,ren2018ran4iqa,pan2018blind,lim2018vr,zhang2018blind,zhang2021uncertainty} are proposed.
Unfortunately, most of the existing IQA methods aim to assess the quality of natural images, while they are limited when dealing with generated images. 

\noindent \textbf{Generative adversarial network assessment.} Several sample based methods have been proposed to assess GAN performance \cite{xu2018empirical}. Among them, Fréchet Inception Distance (FID) \cite{heusel2017gans} is the most popular metric. There are also other proposed metrics like Kernel MMD \cite{gretton2012optimal}, Inception Score (IS) \cite{salimans2016improved}, Mode Score (MS) \cite{che2016mode}, and Wasserstein distance. Note that all these methods measure the deviation between the deep features distribution of generated images and that of real images. Single Image FID \cite{shaham2019singan} aims to compare internal patch statistics difference between generated images and a single reference image, which also focuses on assessing the quality of a generated image distribution. To assess a single generated image, GIQA \cite{gu2020giqa} is proposed as a NR-IQA metric from both learning-based and data-based perspectives. However, it can not evaluate whether the generated image inherits the style of its reference image.


\section{Methodology}

In this section, we first introduce the architecture of our RISA framework. Then we describe how to obtain labeled training data. Finally, our novel objective is presented.

\subsection{Learning-based Framework}
Given a triplet of $\{\{I_{\rm {g}}, I_{\rm {r}}\}, y\}$, where $I_{\rm {g}}$ and $I_{\rm {r}}$ refer to a generated image and its reference image, respectively, and $y \in [0, 1]$ is the target quality score, RISA aims to assess the quality of $I_{\rm {g}}$ according to $I_{\rm {r}}$ as a score $s(I_{\rm {r}}, I_{\rm {g}})$ under the supervision of $y$.
As shown in Figure \ref{fig:pipeline}, RISA consists of a style encoder and multiple binary classifiers.

\noindent \textbf{Style encoder.} Following StarGAN v2 \cite{choi2020stargan}, we implement a convolutional neural network (CNN) with six pre-activation residual blocks \cite{he2016identity} and one fully connected layer as our style encoder $f_{\rm {E}}$. Given an input image $I$, a style encoder learns to extract the style-specific attributes and represents them as the style code $z$. In our pipeline, we first encode $I_{\rm {g}}$ and $I_{\rm {r}}$ into $z_{\rm {g}}$ and $z_{\rm {r}}$ respectively. Then to guarantee the symmetry of RISA, we calculate the absolute value of element-wise subtraction $|z_{\rm g}-z_{\rm r}|$ as the difference of the style codes $z_{\rm {g}}$ and $z_{\rm {r}}$ and use it as the input of multiple binary classifiers.

\noindent \textbf{Multiple binary classifiers.}
A straightforward solution to fit the target quality score $y$ is to adopt a naïve regressor. Unfortunately, empirical results illustrate that a naïve regressor fails to converge in the setting where training images are coarsely annotated (Table \ref{tab:loss-ablation}). This is partially due to the image quality gap within the same iteration. 
To address this issue, inspired by previous works \cite{liu2016age,gu2020giqa}, we train a learning-based network with $K$ binary classifiers instead of a regressor to learn the generated image quality score. To be specific, the $k$-th binary classifier is trained for classifying whether the image quality score (from 0 to 1) is greater than a certain threshold $T_k$, where $ T_k\!=\!(k-1)/K, k\!=\!1,2,\cdots,K$. Denote $p_k$ as the predicted probability of $k$-th binary classifier.  The final predicted score of RISA is the mean of the prediction vector $p(I_{\rm {r}}, I_{\rm {g}})\!=\![p_1,p_2,\cdots,p_K]$  as shown in Figure \ref{fig:pipeline}. 
As a supervision signal of $p(I_{\rm {r}}, I_{\rm {g}})$, the target quality score y is converted to a \emph{binary} label vector $t(I_{\rm {r}},I_{\rm {g}})\!=\![t_1, t_2, \cdots, t_K]$  according to whether it is greater than $T_k$.
For example, if we set $K$ to $5$,  then the target quality score  $y=0.6$ will be converted to $[1, 1, 1, 0, 0]$. In our experiments, we set $K\!=\!16$.

\begin{figure}[t]
    \centering
    \includegraphics[scale=0.4]{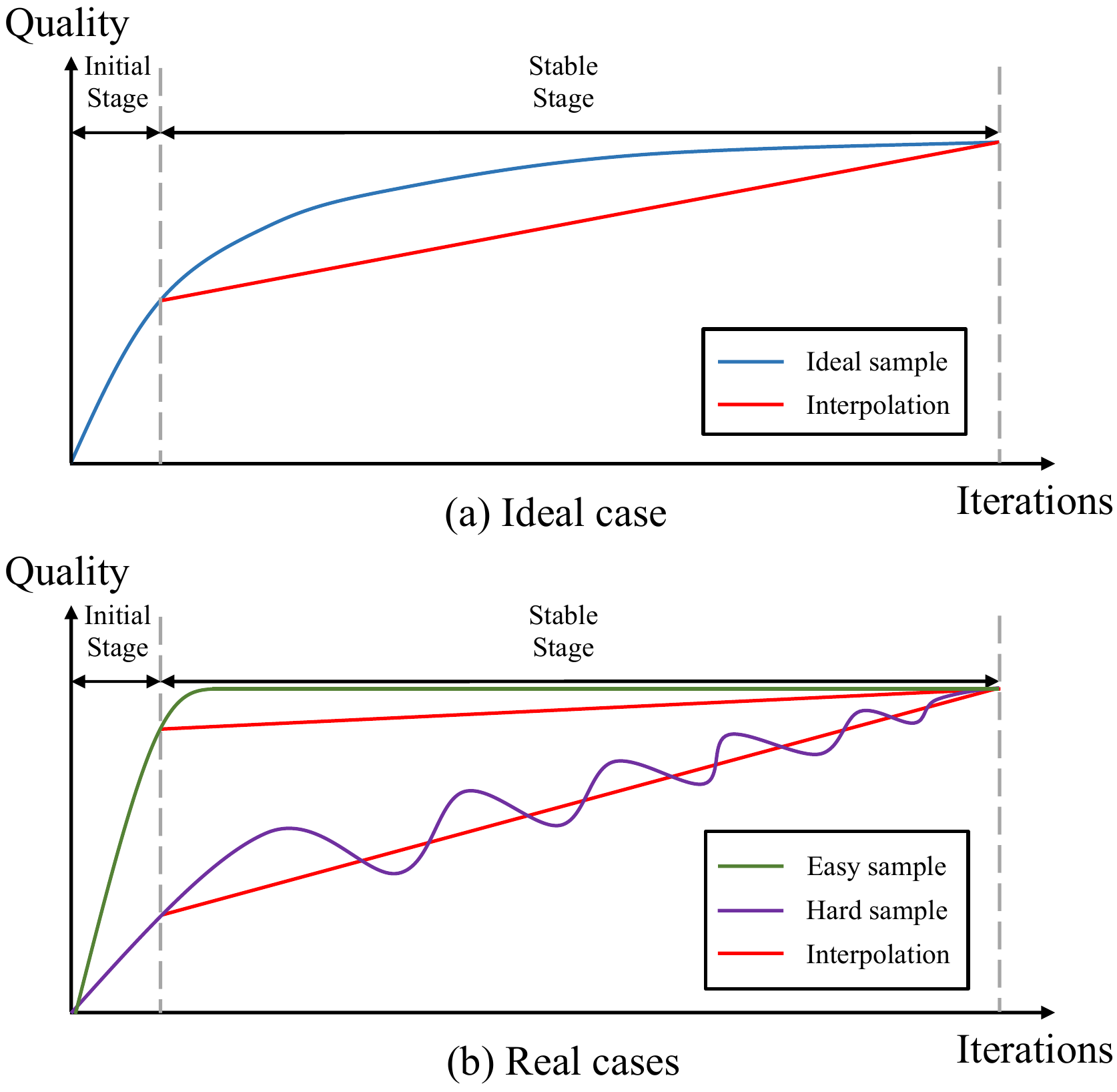}
    \vskip -0.05in
    \caption{An illustration of the image quality variation during the GAN training procedure. (a) shows an ideal case that the vanilla annotation method works well, while (b) indicates two cases that the vanilla annotation method is not suitable. Our proposed pixel-wise interpolation (red lines) can refine the annotations in all these settings.}
    \label{fig:distribution}
    \vskip -0.05in
\end{figure}

\subsection{Data Preparation}
Based on the positive correlation between the quality of generated image and the number of training iterations, we use images generated by intermediate models from GAN training procedure to train RISA and consider the number of training iterations as a weak annotation. To refine this annotation, we propose a pixel-wise interpolation technique.

\noindent \textbf{Coarsely labeled synthesized images.}
Figure \ref{fig:iter} illustrates that the quality of generated images generally evolves with training iterations, in terms of both the visualization effect and FID. Based on this, a vanilla method to obtain training images for RISA is utilizing intermediate models in GAN training to synthesize images. Then each generated image $I_{\rm {g}}^{\rm vanilla}$ is annotated by the number of its corresponding model iterations. To meet the scale of RISA's output, we normalize the annotations into the range $[0, (K-1)/K]$ as $y^{\rm vanilla}$. Here we suppose even synthesized by a converged model, the generated image is still not as perfect as a real image. In our experiments, we only annotate the different views of a same real image with the highest quality score $1$, where the views are produced via style-preserving augmentation, such as scaling, cropping and clipping. 

This vanilla method is reasonable for ideal cases as Figure \ref{fig:distribution}(a), where the quality of the generated images monotonically increases with iterations. However, in real cases, it is unsuitable. As illustrated in Figure \ref{fig:distribution}(b), the whole GAN training process could be separated into two successive stages, namely the initial stage and the stable stage. Empirically, we recommend the elbow point of FID curve as an appropriate stage boundary. During the initial stage, the quality of generated images all improves rapidly as ideal cases. However,  during the stable stage, the quality of easy samples becomes stable and invariant after a few iterations, and the quality of hard samples presents an oscillatory convergence. As a result, the number of model iterations can not represent the image quality during the stable stage.

\noindent \textbf{Pixel-wise interpolation.} To tackle this problem, we introduce a pixel-wise interpolation technique (red lines in Figure \ref{fig:distribution}) as an estimation approach to capture the quality changes during the stable stage. Given a pair of source image and reference image, we observe that the image generated by an intermediate model with iterations around the stage boundary have lower quality than the image synthesized by a finally converged model at the end of the whole GAN training procedure. For simplicity, $\{\{I_{\rm {g}}^{\rm low}, I_{\rm {r}}\}, y^{\rm low}\}$ refers to the former, and $\{\{I_{\rm {g}}^{\rm high}, I_{\rm {r}}\}, y^{\rm high}\}$ denotes the latter. To produce images with quality between $I_{\rm {g}}^{\rm low}$ and $I_{\rm {g}}^{\rm high}$, we implement linear interpolation in the pixel space:
\begin{equation}
    I_{\rm {g}}^{\rm inter} = \epsilon I_{\rm {g}}^{\rm high} + (1 - \epsilon) I_{\rm {g}}^{\rm low}, y^{\rm inter} = \epsilon y^{\rm high} + (1 - \epsilon) y^{\rm low},
\end{equation}
where $I_{\rm {g}}^{\rm inter}$ and $y^{\rm inter}$ represent the interpolated image and its quality score, respectively. $\epsilon \in (0, 1)$ is an interpolation factor. By varying $\epsilon$, we could generate a series of images with different quality between the quality of $I_{\rm {g}}^{\rm low}$ and $I_{\rm {g}}^{\rm high}$.

A natural question to ask is why pixel-wise interpolation is effective to generate images of different quality. As illustrated in Figure \ref{fig:iter}, during the stable stage, generated images could all preserve the content of their source images perfectly and maintain the style of their reference images generally. The model mainly focuses on improving the generation of detailed textures. Pixel-wise interpolation with different $\epsilon$ could estimate these local texture variations while have no influence on the global structure and texture. Empirical results in Table \ref{tab:inter-ablation} also indicate the performance improvements gained from the pixel-wise interpolation technique.

As a summary, generated images in RISA's training data are synthesized as:
\begin{equation}
 I_{\rm {g}} = \begin{cases}
 I_{\rm {g}}^{\rm vanilla}, & \rm{during\ the\ initial\ stage}; \\
 I_{\rm {g}}^{\rm inter}, & \rm {during\ the\ stable\ stage}.
 \end{cases}
\end{equation}

\begin{table*}[t]
\centering 
\begin{tabular}{lcc|c|cc}
\toprule
          & CelebA-HQ    & AFHQ & Yosemite  & Church & Bedroom    \\ \cline{2-6}
          & \multicolumn{2}{c|}{StarGAN v2}    & \multicolumn{1}{c|}{MSGAN}           & \multicolumn{2}{c}{Swap Autoencoder}  \\ \hline
NIQE      & 60.09$\pm${0.99}\%  & 60.12$\pm${0.97}\%  & 52.18$\pm${1.39}\% & 54.48$\pm${2.39}\%  & 54.37$\pm${6.01}\%    \\
Deep-IQA  & 50.22$\pm${4.66}\%  & 60.57$\pm${3.55}\%  & 53.27$\pm${0.58}\% & 54.17$\pm${5.69}\%  & 50.32$\pm${4.34}\%    \\
NIMA      & 52.39$\pm${0.65}\%  & 58.17$\pm${3.68}\%  & 47.82$\pm${3.33}\% & 54.18$\pm${2.82}\%  & 48.00$\pm${2.27}\%    \\
GMM-GIQA  & 52.25$\pm${1.79}\%  & 68.82$\pm${2.52}\%  & 57.09$\pm${1.14}\% & 70.95$\pm${2.35}\%  & 64.42$\pm${4.27}\%    \\
KNN-GIQA  & 50.94$\pm${7.25}\%  & 70.16$\pm${1.84}\%  & 49.82$\pm${1.48}\% & 50.90$\pm${2.57}\%  & 49.67$\pm${1.18}\%    \\ \hline
SSIM      & 55.01$\pm${1.60}\%  & 57.57$\pm${3.25}\%  & 45.45$\pm${3.61}\% & 64.67$\pm${5.24}\%  & 77.51$\pm${4.11}\%    \\
MS-SSIM   & 52.98$\pm${1.01}\%  & 61.77$\pm${5.52}\%  & 44.37$\pm${1.09}\% & 60.78$\pm${1.62}\%  & 67.13$\pm${5.39}\%    \\
PSNR      & 51.81$\pm${2.52}\%  & 55.47$\pm${2.36}\%  & 44.73$\pm${3.58}\% & 65.56$\pm${2.37}\%  & 77.53$\pm${3.82}\%    \\ 
FSIM      & 56.75$\pm${4.58}\%  & 61.17$\pm${5.52}\%  & 51.09$\pm${0.59}\% & 58.68$\pm${2.01}\%  & 56.38$\pm${2.28}\%    \\
LPIPS     & 54.86$\pm${1.18}\%  & 75.41$\pm${2.02}\%  & 57.45$\pm${0.99}\% & 66.77$\pm${1.48}\%  & 71.15$\pm${1.57}\%    \\ 
SIFID     & 57.76$\pm${1.55}\%  & 69.27$\pm${1.55}\%  & 58.55$\pm${3.42}\% & 69.18$\pm${3.92}\%  & 72.81$\pm${1.75}\%    \\ \hline
\textbf{RISA(ours)}  & \textbf{70.54$\pm${1.84}\%} & \textbf{80.36$\pm${1.44}\%} & \textbf{64.36$\pm$1.79\%} &  \textbf{73.96$\pm$2.83\%} & \textbf{83.55$\pm$3.37\%} \\ \bottomrule
\end{tabular}
\vskip -0.05in
\caption{Consistency (in \%) with human judgments (50\% indicates a random guess). The training data for RISA and the samples for human judgments are generated by the models with the same architecture.}
\vskip -0.15in
\label{tab:intra-model}
\end{table*}

\begin{figure}[t]
    \centering
    \includegraphics[scale=0.34]{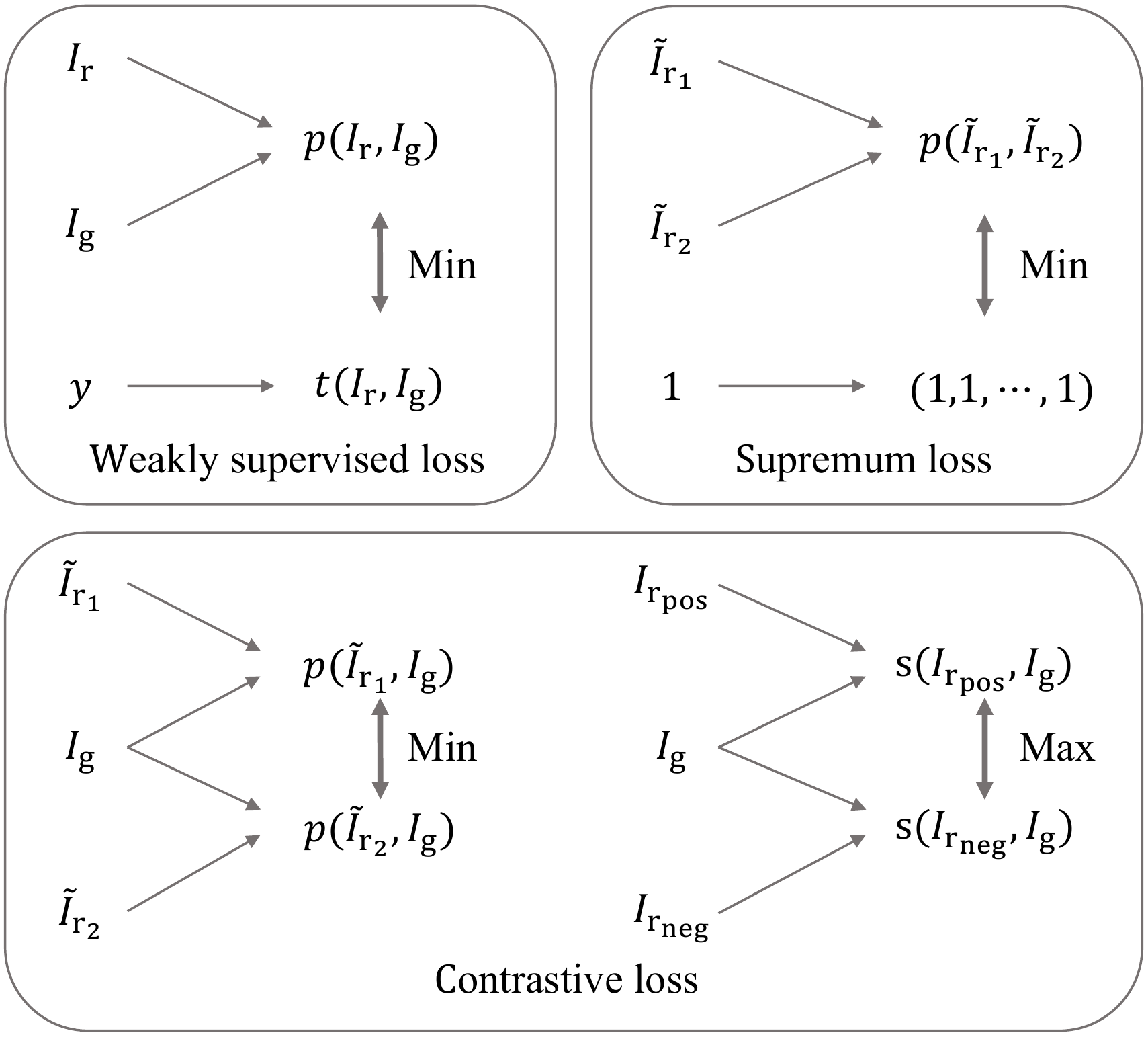}
    \vskip -0.05in
    \caption{RISA's training objective. It contains a weakly supervised loss, a contrastive loss and a supremum loss.}
    \label{fig:obj}
    \vskip -0.15in
\end{figure}

\subsection{Training Objective}

As illustrated in Figure \ref{fig:obj}, our training objective includes three terms: 1) a weakly supervised loss to learn the \emph{pseudo} quality score, 2) an unsupervised contrastive loss to effectively capture the style similarity between a generated image and its reference image, and 3) a supremum loss to learn the style consistency of two augmented views from a real image.

\noindent \textbf{Weakly supervised loss.}
We implement the weakly supervised loss term utilizing the binary cross entropy:
\begin{equation}
\begin{aligned}
& \mathcal{L}_{\rm sup}(p({I}_{\rm {r}}, I_{\rm {g}}), t({I}_{\rm {r}}, I_{\rm {g}}))\\
=& \sum\nolimits_{k=1}^{K} (t_k \log p_k + (1-t_k) \log(1-p_k)),
\end{aligned}
\end{equation}
where $p({I}_{\rm {r}}, I_{\rm {g}})$ and $t({I}_{\rm {r}}, I_{\rm {g}})$ refer to the prediction vector and the binary label vector, respectively, and their index is ignored for simplicity (the same below). 

\noindent \textbf{Contrastive loss.}
 We employ the contrastive loss to capture the style similarity between a generated image and its reference image, which is essential to the generalization ability of our framework. In specific, we produce two views $\widetilde{I}_{\rm {r}_1}, \widetilde{I}_{\rm {r}_2}$ of reference image $I_{\rm {r}}$ via data augmentation. To preserve \emph{style-relevant} information, the augmentation operations only consist of scaling, cropping and clipping. Then they are fed to our framework for producing the prediction vectors $p(\widetilde{I}_{\rm {r}_1},I_{\rm {g}}), p(\widetilde{I}_{\rm {r}_2},I_{\rm {g}})$ and the prediction scores $s(\widetilde{I}_{\rm {r}_1},I_{\rm {g}}), s(\widetilde{I}_{\rm {r}_2},I_{\rm {g}})$.
 We treat $p(\widetilde{I}_{\rm {r}_1},I_{\rm {g}})$ and $p(\widetilde{I}_{\rm {r}_2},I_{\rm {g}})$ as a positive pair and minimize the distance of them. Thus the positive part of constrastive loss is expressed as:
\begin{equation}
\begin{aligned}
\mathcal{L}_{\rm pos}(p(\widetilde{I}_{\rm {r}_1},I_{\rm {g}}), p(\widetilde{I}_{\rm {r}_2},I_{\rm {g}}))=\Vert p(\widetilde{I}_{\rm {r}_1},I_{\rm {g}}) - p(\widetilde{I}_{\rm {r}_2},I_{\rm {g}}) \Vert^2_2, 
\end{aligned}
\end{equation}
where $\Vert \cdot \Vert^2_2$ denotes the squared Euclidean norm. 

In addition, we consider $s(I_{\rm {r_{neg}}}, I_{\rm {g}})$ and $s(I_{\rm {r_{pos}}}, I_{\rm {g}})$ as a negative sample and a positive sample, respectively, where $I_{\rm {r}_{\rm {pos}}}$ could be $\widetilde{I}_{\rm {r}_1}$ or $\widetilde{I}_{\rm {r}_2}$ and $I_{\rm {r_{neg}}}$ refers to a randomly selected reference image. To enlarge the difference between the negative sample and the positive sample, the negative part of the contrastive loss is defined as:
\begin{equation}
\begin{aligned}
&\mathcal{L}_{\rm neg}(s(I_{\rm {r_{neg}}}, I_{\rm {g}}), s(I_{\rm {r_{pos}}}, I_{\rm {g}})) \\
=& {\rm max}(0, s(I_{\rm {r_{neg}}}, I_{\rm {g}}) - \gamma s(I_{\rm {r_{pos}}}, I_{\rm {g}})),
\end{aligned}
\end{equation}
where $\gamma$ is a penalty factor to control the restraint strength. We set $\gamma\!=\!0.5$ in our experiments.

\noindent \textbf{Supremum loss} is introduced as a supplement of the weakly supervised loss in which the pseudo quality score is strictly less than $1$. Since different views of a reference image in contrastive loss preserve the style-relevant information, the quality score of $\widetilde{I}_{\rm {r}_1}$ and $\widetilde{I}_{\rm {r}_2}$ should be the highest score 1. Therefore, as an additional supervision signal, a supremum loss is represented as: 
\begin{equation}
\mathcal{L}_{\rm supre}(p(\widetilde{I}_{\rm {r}_1}, \widetilde{I}_{\rm {r}_2}), \mathbf{1}) = \sum\nolimits_{k=1}^{K} \log p_k,
\end{equation}
where $\mathbf{1}$ denotes a vector of ones with length $K$. 

\noindent \textbf{Full objective.}
To sum up, our full objective is given as:
\begin{equation}
\begin{aligned}
    \mathcal{L}=&\mathcal{L}_{\rm sup}(p({I}_{\rm r}, I_{\rm {g}}), t({I}_{\rm {r}}, I_{\rm {g}})) \\
    &+ \lambda_{\rm p}\mathcal{L}_{\rm pos}(p(\widetilde{I}_{\rm {r}_1},I_{\rm {g}}), p(\widetilde{I}_{\rm {r}_2},I_{\rm {g}})) \\
    & +\lambda_{\rm n}\mathcal{L}_{\rm neg}(s(I_{\rm r_{\rm neg}}, I_{\rm g}), s(I_{\rm {r}_{\rm pos}}, I_{\rm g})) \\
    &+ \lambda_{\rm s}\mathcal{L}_{\rm supre}(p(\widetilde{I}_{\rm {r}_1}, \widetilde{I}_{\rm {r}_2}), \mathbf{1}). \\ 
\end{aligned}
\end{equation}
In our experiments, we set $\lambda_{\rm p}$, $\lambda_{\rm n}$ and $\lambda_{\rm s}$ to $1$ for simplicity.


\section{Experiments}
\label{sec:expr}

In this section, we start by introducing the experimental setup. Then extensive results are reported to demonstrate the effectiveness and generalization ability of RISA. In addition, we carefully analyze each components of RISA and compare different training configurations as ablation studies. 

\subsection{Experimental Setup}

\noindent \textbf{Datasets and genarative models.}
We conduct our experiments on five datasets: Yosemite \cite{zhu2017unpaired}, CelebA-HQ \cite{karras2017progressive}, AFHQ \cite{choi2020stargan},  LSUN Church and Bedroom \cite{yu15lsun}, all at the resolution of $256\times256$. For multi-domain GAN training, we separate Yosemite into two domains of summer and winter, CelebA-HQ into two domains of male and female and AFHQ into three domains of cat, dog, and wildlife.
 
Following the original papers, we train DRIT \cite{lee2018diverse} and MSGAN \cite{huang2018multimodal} on Yosemite and CelebA-HQ, StarGAN v2 \cite{choi2020stargan} on CelebA-HQ and AFHQ, and Swap Autoencoder (Swap AE) \cite{park2020swapping} on CelebA-HQ, LSUN Church and Bedroom. 

\begin{table*}[t]
\centering 
\small

\begin{tabular}{lcccc|cc}
\toprule 
   & \multicolumn{4}{c|}{CelebA-HQ}         & \multicolumn{2}{c}{Yosemite}      \\ \cline{2-7}
   
                                      & DRIT               &      MSGAN         &    MSGAN-Swap AE   & StarGAN v2-Swap AE  &        DRIT          &  DRIT-MSGAN           \\ \hline
                                                                                                                            
NIQE     & 50.59$\pm${3.05}\% & 48.77$\pm${4.44}\% & 53.46$\pm${4.01}\% & 46.97$\pm${4.69}\%  &  51.18$\pm${0.20}\%  &  55.13$\pm${3.54}\%   \\
Deep-IQA & 50.58$\pm${1.43}\% & 46.60$\pm${4.37}\% & 41.51$\pm${0.89}\% & 51.14$\pm${2.90}\%  &  45.25$\pm${3.15}\%  &  53.53$\pm${5.51}\%   \\
NIMA     & 49.42$\pm${3.02}\% & 54.14$\pm${1.63}\% & 58.28$\pm${1.19}\% & 47.73$\pm${3.80}\%  &  52.35$\pm${5.03}\%  &  54.17$\pm${9.43}\%   \\
GMM-GIQA & 56.93$\pm${2.21}\% & 59.49$\pm${5.23}\% & 58.07$\pm${4.00}\% & 49.81$\pm${1.87}\%  &  60.96$\pm${3.01}\%  &  60.90$\pm${5.94}\%   \\
KNN-GIQA & 51.46$\pm${4.48}\% & 51.47$\pm${0.79}\% & 53.67$\pm${2.14}\% & 52.65$\pm${2.19}\%  &  51.17$\pm${2.82}\%  &  50.64$\pm${4.53}\%   \\ \hline
SSIM     & 49.15$\pm${3.68}\% & 52.93$\pm${1.30}\% & 58.49$\pm${2.24}\% & 55.49$\pm${5.11}\%  &  51.49$\pm${2.06}\%  &  48.40$\pm${2.97}\%   \\
MS-SSIM  & 48.26$\pm${5.23}\% & 56.35$\pm${1.71}\% & 62.26$\pm${2.24}\% & 59.47$\pm${3.60}\%  &  50.29$\pm${0.21}\%  &  45.19$\pm${2.08}\%   \\
PSNR     & 50.58$\pm${2.36}\% & 57.31$\pm${6.02}\% & 66.04$\pm${4.39}\% & 57.01$\pm${2.56}\%  &  49.39$\pm${4.63}\%  &  47.44$\pm${1.20}\%   \\ 
FSIM     & 45.37$\pm${3.56}\% & 58.29$\pm${0.73}\% & 58.70$\pm${3.14}\% & 53.22$\pm${6.05}\%  &  47.32$\pm${3.16}\%  &  54.49$\pm${2.76}\%   \\
LPIPS    & 56.65$\pm${0.18}\% & 64.15$\pm${1.79}\% & 67.30$\pm${4.62}\% & 60.04$\pm${1.49}\%  &  64.49$\pm${0.86}\%  &  57.37$\pm${0.45}\%   \\ 
SIFID    & 55.78$\pm${1.94}\% & 62.43$\pm${3.49}\% & 62.89$\pm${1.36}\% & 57.77$\pm${4.51}\%  &  64.51$\pm${4.39}\%  &  55.13$\pm${4.73}\%   \\ \hline
\textbf{RISA(ours)}  & \textbf{65.32$\pm$4.26\%}  &  \textbf{70.73$\pm$1.20\%}  &\textbf{72.75$\pm$0.59\%}& \textbf{68.94$\pm$1.63\%}  &  \textbf{67.45$\pm$2.66\%}  & \textbf{63.46$\pm$2.36\%} \\ \bottomrule
\end{tabular}
\vskip -0.05in
\caption{Cross-model consistency (in \%) with human judgments. "Cross-model" indicates that the training data for RISA and the samples for human judgments are generated by the models with differenet architectures. To be specific, for CelebA-HQ and Yosemite, the generative models to synthesis RISA's training images are StarGAN v2 and MSGAN, respectively. The generative models used for human judgments are either intermediate models at different training iterations during the stable stage, e.g., DRIT, or two converged models with different architectures, e.g., MSGAN-Swap AE.}
\vskip -0.15in
\label{tab:inter-model}
\end{table*}

\noindent \textbf{Implementation details.}
For each generative model, we first choose 7 intermediate models (DRIT and MSGAN trained for 1, 10, 20, 40, 80, 200 and 1200 epochs, StarGAN v2 trained for 1k, 2k, 4k, 6k, 8k, 10k and 100k iterations and Swapping Autoencoder trained for 100k, 200k, 500k, 1M, 2M, 5M and 25M images), each of which is utilized to synthesize 1k images. Then we execute the pixel-wise interpolation using the last two models mentioned above, e.g., models at 10k and 100k iterations for StarGAN v2, with $\epsilon=0.1, 0.2, \cdots, 0.9$. We synthesize 1k interpolated images under each $\epsilon$. Finally, for each dataset, 16k images with 16 different quality scores ($k/16, k=1, 2, \cdots, 15$) is obtained as the training images.

RISA is implemented in PyTorch \cite{paszke2019pytorch}. The batch size is set to 4 and the model is trained for 100 epochs using a single NVIDIA RTX 2080Ti GPU. We use the Adam \cite{kingma2014adam} with $\beta_1=0$ and $\beta_2=0.99$. The weight decay and the learning rate are set to $10^{-4}$. The weights of all modules are initialized using He initialization \cite{he2015delving} and all bias are set to zero.

\noindent \textbf{Human evaluation.} 
To compare the effectiveness of different metrics, we test the consistency of each metric with human judgments through various binary classification experiments. In specific, each testing sample is a triplet consisting of a \emph{reference} image and two \emph{generated} images synthesized by different generative models. Human observers are required to independently select the generated image of higher quality according to the reference image. To guarantee the experiments are nontrivial, the generative models are either two intermediate models during the stable stage (Table \ref{tab:intra-model}, \ref{tab:inter-model}) or directly two converged models with different architectures (Table \ref{tab:inter-model}). Samples that all observers reach a consensus on (about 400 samples per setting) are used to evaluate metrics. We report the average consistency and the standard deviations by dividing samples into 3 equal parts. 

\noindent \textbf{Baselines.}
We compare our methods with 11 baselines:\\
\textbf{NR-IQA methods}: NIQE \cite{mittal2012making} calculates the distance between the multivariate Gaussian model of the test image and a natural scene statistic model as a quality measure. Deep-IQA \cite{bosse2016deep} uses a deep network to assess the quality of various patches randomly sampled from the test image. NIMA \cite{talebi2018nima} assesses the quality of images using a CNN trained with massive labeled images. GIQA \cite{gu2020giqa} assesses an image from both learning-based and data-based perspectives. The recommended GMM-GIQA and KNN-GIQA are adopted in our experiments.\\
\textbf{FR-IQA methods}: 
SSIM \cite{wang2004image} measures the discrepancy of two images' luminance, contrast and structure. MS-SSIM \cite{wang2003multiscale} calculates the image's SSIM in multiple scales. PSNR \cite{huynh2008scope} considers the ratio between the maximum possible power of a signal and the power of corrupting noise. FSIM \cite{zhang2011fsim} is a variation of SSIM, using different weights to represent the importance of different regions in image. LPIPS \cite{zhang2018unreasonable} trains evaluation networks using 3 methods (named \emph{lin}, \emph{tune} and \emph{scratch}). SIFID \cite{shaham2019singan} applies FID by viewing features of a image as a distribution.

\subsection{Results}

\begin{figure}[t]
    \centering
    \includegraphics[scale=0.23]{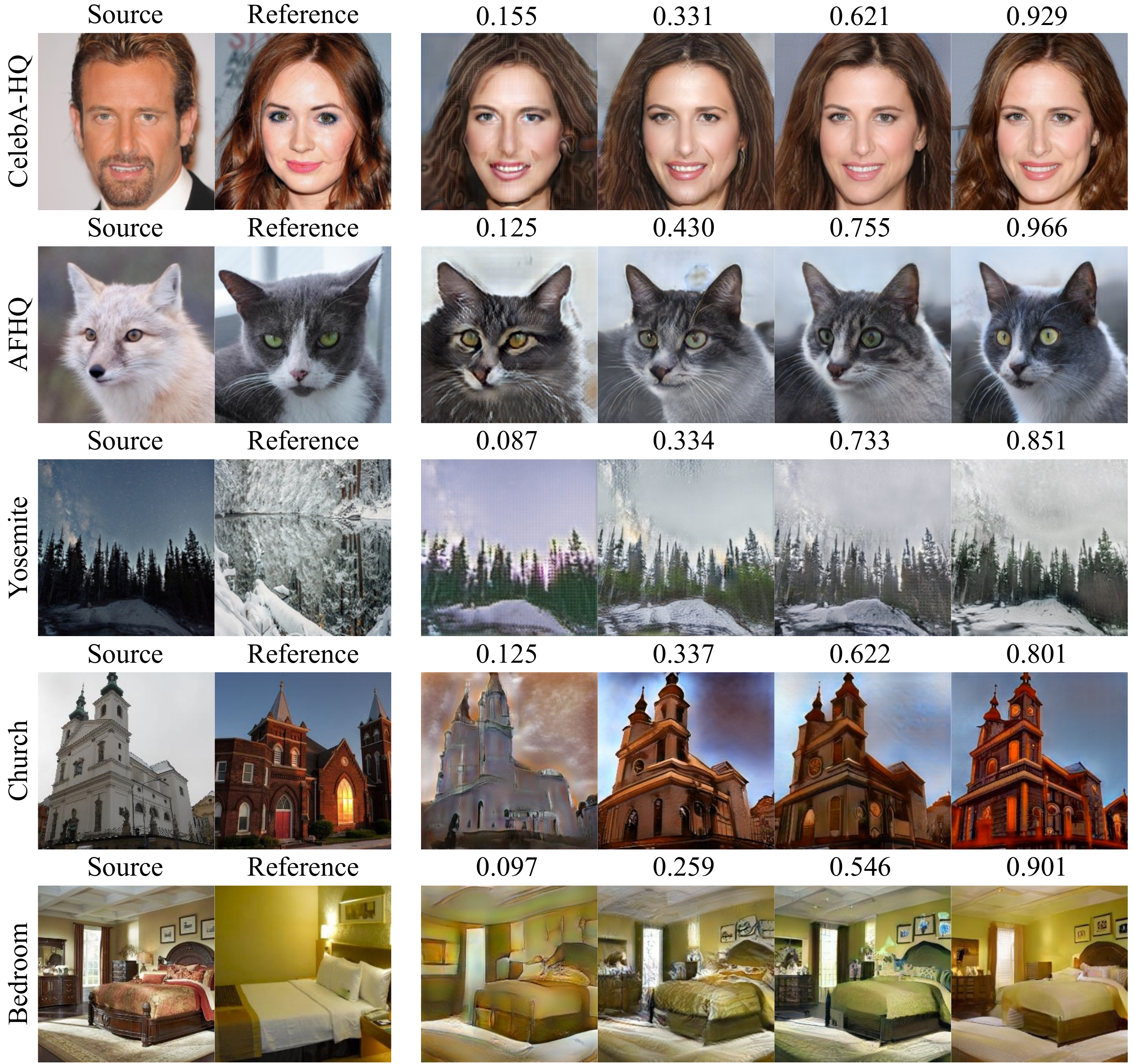}
    \vskip -0.05in
    \caption{A quick evaluation on images with visible quality gaps. The left two columns are the source and reference images, and the right four columns show images generated by different models. The value above each generated image is the quality score assessed by RISA.}
    \label{fig:quick_eval}
    \vskip -0.2in
\end{figure}

\noindent \textbf{A quick evaluation.}
We conduct an intuitive evaluation to verify the effectiveness of RISA. In particular, for each dataset, we manually select a series of images with visible quality gaps and utilize RISA to assess them. Empirical results in Figure \ref{fig:quick_eval} illustrate that images of higher quality can get a higher score via RISA, and vice versa.

\begin{figure}[t]
    \centering
    \includegraphics[scale=0.3]{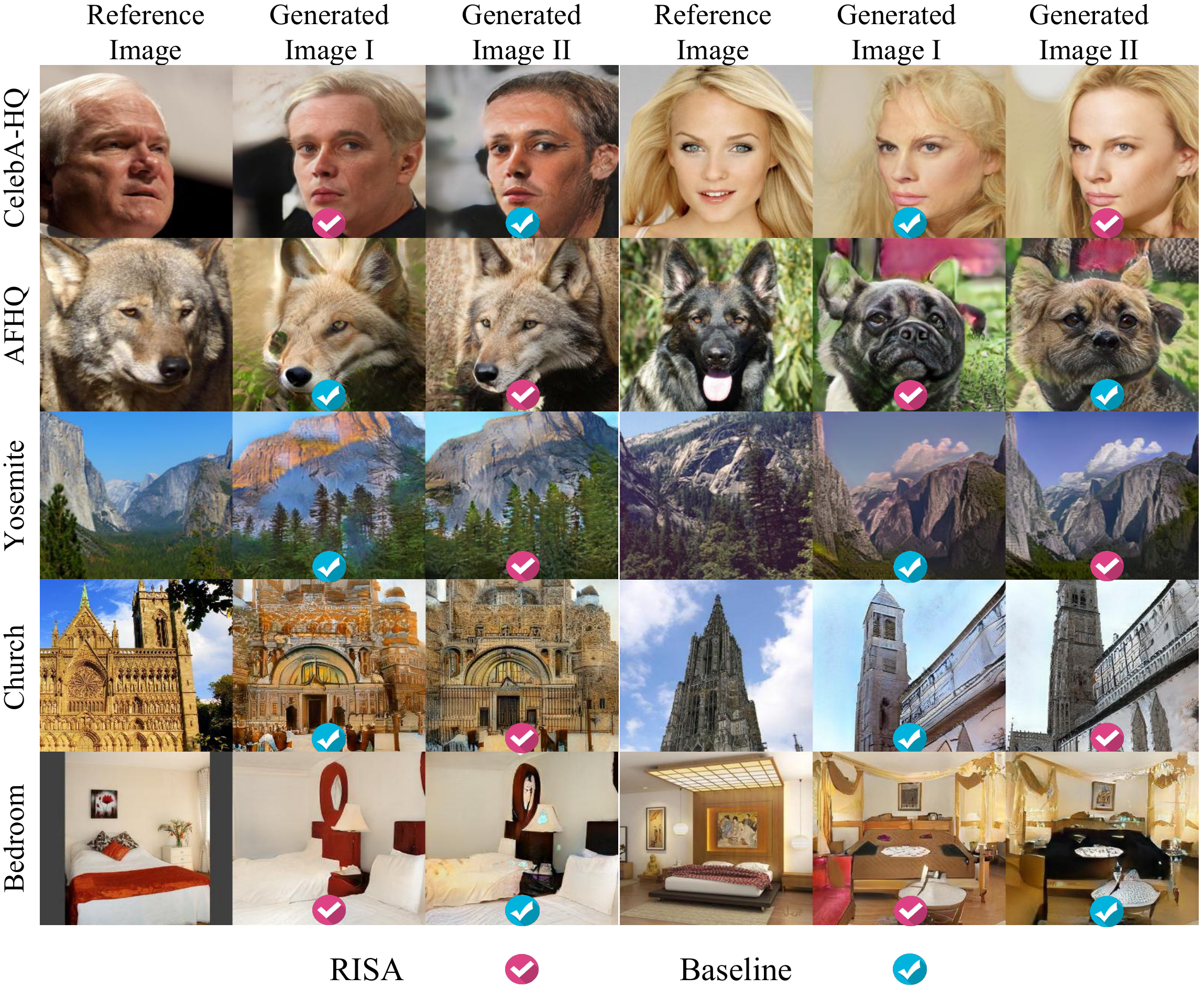}
    \vskip -0.05in
    \caption{Compared with the most competitive baseline on each dataset (NIQE for CelebA-HQ, LPIPS for AFHQ, SIFID for Yosemite, GMM-GIQA for Church, and PSNR for Bedroom), RISA achieves better performance on selecting the generated image with higher quality.}
    \label{fig:compare}
    \vskip -0.15in
\end{figure}

\noindent \textbf{Performance comparisons.}
Table \ref{tab:intra-model} demonstrates the consistency of metrics with human judgments on various datasets and generative models, and we highlight the best performance in bold. From the results, our proposed RISA is more consistent with human judgments than other image assessment methods. Visualizations in Figure \ref{fig:compare} also indicate that our proposed method could assess the quality of a generated image according  to its reference image.

In addition, we report the favorable generalization performance of RISA in Table \ref{tab:inter-model}, where the RISA's training images and the human judgement samples are synthesized by generative models with different architectures. Moreover, the settings in Table \ref{tab:inter-model} can be further divided into two categories, according to the samples are generated by multiple intermediate models (single-model settings) or synthesized by two different converged models (double-model settings). Compared with baselines, results in single-model settings indicate that RISA transfers well across different models. Furthermore, results in double-model settings demonstrate that RISA effectively chooses the image with higher style similarity to its reference image.

For baselines, the assessment of NR-IQA methods only leverage the generated image. They are not able to evaluate the image's style similarity to its reference image. Although FR-IQA methods calculate the discrepancy between the an image and its reference image, empirical results indicate that they are less suitable to measure the style similarity.

\subsection{Ablation Study}

\begin{table}[t]
\centering 
\small

\begin{tabular}{lcc}
\toprule
    & \multicolumn{2}{c}{CelebA-HQ} \\ \cline{2-3}                              & StarGAN v2  & StarGAN v2-Swap AE \\ \hline
Naïve Regressor                   & N/A      & N/A  \\
+ Multi-Classifiers               & 67.78$\pm${2.84}\%      & 53.22$\pm${2.98}\%  \\
+ Contrastive Loss                & 69.52$\pm${1.00}\%      & 61.36$\pm${2.82}\%  \\
+ Supremum Loss                   & \textbf{70.54$\pm${2.47}\%}   & \textbf{68.94$\pm${1.93}\%}       \\ \bottomrule
\end{tabular}
\vskip -0.05in
\caption{ Consistency (in \%) with human judgments corresponding to various components. The generative model utilized to build training set for RISA is StarGAN v2.}
\vskip -0.15in
\label{tab:loss-ablation}
\end{table}

\noindent \textbf{Effectiveness of various components.}
We examine each individual component in our framework in Table \ref{tab:loss-ablation}, where each component is cumulatively added to a naïve regressor. We find that only with a naïve regressor, RISA fails to converge and gives the same prediction for all samples because the quality labels of training images are too coarse.
In the "StarGAN v2" setting, it is intriguing that simply applying multiple binary classifiers instead of a naïve regressor makes RISA achieve a competitive performance compared with the standard setting. It demonstrates that the multiple binary classifiers effectively suppress the label noise.  
In the "StarGAN v2-Swap AE" setting, since the two generated images are synthesized by two powerful converged models, the style similarity is a more important aspect to assess. We can find that the contrastive loss is critical to capture the style similarity. 
In addition, using the supremum loss also lead RISA to achieve higher consistency with human preference.

\noindent \textbf{Effectiveness of pixel-wise interpolation.}
Table \ref{tab:inter-ablation} compares the performance of different ways to build the training set for RISA. 
As the samples used for human judgments are generated by intermediate models during the stable stage, it is natural to compare the setting that the training images are totally generated using models during the stable stage (Stable Stage Only) with the standard setting. To verify the effectiveness of pixel-wise interpolation, we also conduct an experiment which is trained on images generated by models from both the initial stage and the stable stage (+ Initial Stage). Empirical results show that combining images of low quality and of high quality promotes RISA to assess more accurately and refining the coarse labels with pixel-wise interpolation further improves RISA's performance.

\begin{table}[t]
\centering 
\small

\begin{tabular}{lcc}
\toprule
                 & \multicolumn{2}{c}{CelebA-HQ} \\ \cline{2-3}  
                & StarGAN v2  & StarGAN v2-Swap AE \\ \hline
Stable Stage Only    & 61.10$\pm${4.45}\%      & 60.42$\pm${2.19}\%    \\
+ Initial Stage     & 66.04$\pm${3.45}\%      & 63.07$\pm${1.23}\% \\
+ Interpolation    &\textbf{70.54$\pm${2.47}\%}     & \textbf{68.94$\pm${1.93}\%}  \\ \bottomrule
\end{tabular}
\vskip -0.05in
\caption{Consistency (in \%) with human judgments corresponding to different ways of building the training set. StarGAN v2 is utilized to build the training set for RISA.}
\label{tab:inter-ablation}

\end{table}

\subsection{Discussion}

 Assessing a single reference-guided synthesized image is an area of great practical significance but lacks of the research. Although our novel RISA framework performs well on various datasets and settings, there are still opening problems need to be further explored. RISA regards the generated image's style similarity to its reference image as a more significant component to assess, while ignores the content similarity to its source image. Although trained with few iterations, a generative model can synthesize the image which maintains the content (or spatial structure) of its source image perfectly as shown in  Figure \ref{fig:iter}. In contrast, the style-relevant textures are continuously evolving through the whole training procedure. In addition, prior works \cite{gatys2015texture, gatys2016image, huang2017arbitrary, huang2018multimodal} also support our opinion since they mainly focus on designing effective objectives to render the style.

\section{Conclusion}


In this paper, we propose RISA, a learning-based framework to assess a single reference-guided synthesized image.
Notably, RISA works in a weakly supervised scheme without any human annotations. In specific, the training images are generated by intermediate models in RIS training and the corresponding labels are annotated by the number of models' iterations. To suppress the label noise, we propose a pixel-wise interpolation technique and adopt multiple binary classifiers. Moreover, an unsupervised contrastive loss is introduced to effectively capture the style similarity. Compared with existing single image assessment metrics, RISA achieves higher consistency with human preference on various datasets and transfers well across models. We believe that our work will contribute to improving user experience in real-world RIS applications and motivate future researches on developing more effective assessment metrics.



\section{Acknowledgements}
This work is supported in part by the National Science and Technology Major Project of the Ministry of Science and Technology of China under Grant 2018AAA0100701, the National Natural Science Foundation of China under Grants 61906106 and 62022048, and Beijing Academy of Artificial Intelligence (BAAI).

\bibliography{reference.bib}
\end{document}